\documentclass[11pt]{article}

\usepackage[T1]{fontenc}
\usepackage[utf8]{inputenc}
\usepackage[margin=1in]{geometry}
\usepackage{amsmath,amssymb,amsfonts}
\usepackage{booktabs}
\usepackage{float}
\usepackage{array}
\usepackage{longtable}
\usepackage{enumitem}
\usepackage{natbib}
\usepackage{hyperref}
\usepackage{url}

\title{Do Transaction-Level and Actor-Level AML Queues Agree?\\
An Empirical Evaluation of Granularity Effects on the Elliptic++ Graph}

\author{Ankur Malik\\
Hudson Data, New York, NY, USA\\
\texttt{ankur@hudsondata.com}\\
ORCID: 0009-0005-0676-6239}

\date{April 2026}

\begin{document}

\maketitle

\begin{abstract}
Graph-based anti-money laundering (AML) systems on blockchain networks can score suspicious activity at two natural granularity levels---individual transactions or actor addresses---yet the compliance action is conducted per actor. This paper contributes a data science evaluation methodology for measuring how the choice of scoring granularity affects investigation queue composition under fixed review budgets. We formalize the evaluation through a projection framework that maps transaction-level scores to the actor-level action unit via four aggregation operators, and introduce budgeted investigation metrics---yield@budget, burden decomposition, and case fragmentation---that capture operational consequences standard classification metrics miss. Using the public Elliptic++ Bitcoin dataset (203,769 transactions; 822,942 address occurrences), we train independent random forest classifiers at each granularity level under a causal temporal protocol and compare the resulting review queues through Jaccard overlap, burden decomposition (illicit, known-licit, and unknown cases per 100 reviews), and feature-matching ablations. At a one-percent review budget, temporal evaluation across ten test timesteps yields a mean Jaccard of 0.374 (standard deviation 0.171); static pooled evaluation yields 0.087 (95\% confidence interval [0.079, 0.094]). An enriched address model receiving all 237 features aggregated from transaction data produces even lower overlap (Jaccard = 0.051), with 4.3\% illicit per 100 reviews versus 30.2\% for the transaction-projected queue. Address-level detection value is temporally concentrated: two timesteps exceed 91\% illicit per 100 reviews while the static address-queue burden is only 3.4\%. A fixed hybrid consensus policy underperforms the best single-level queue by 5.05 percentage points (confidence interval [$-$10.2pp, $-$0.9pp]), indicating that naive score combination cannot exploit temporally heterogeneous disagreement. These findings establish that, on this dataset and under the primary noisy-OR projection, scoring granularity is a consequential design variable for AML investigation systems---same data, same budget, different queues, different addresses investigated. The evaluation methodology generalizes to any detection domain where scores produced at one granularity must be compared at another.
\end{abstract}


\section{Introduction}

Anti-money laundering compliance teams operate under fixed review budgets. A scored queue shapes which addresses are investigated and, by exclusion, which are not. If transaction-level and actor-level detection models produce different queues from the same underlying data, then the choice of scoring granularity---not just classifier accuracy---can materially change who gets investigated.

Graph-based AML detection on blockchain networks operates at two natural levels of granularity. Transaction-level models score individual transfers between addresses using local and topological features. Address-level models score blockchain addresses using account-level attributes and behavioral summaries. Both approaches have been studied independently, with evaluation focused on standard classification metrics: AUROC, F1, precision-recall. Yet the compliance action unit is the actor. An analyst reviews an address (or the entity behind it), not a single transaction. When transaction-level scores must be aggregated to the actor level---via projection operators such as noisy-OR, max-score, or mean---the resulting review queue may diverge from one constructed natively at the actor level.

This divergence has operational consequences that standard classification metrics do not capture. Two models with identical AUROC can produce substantially different top-$K$ queues, directing investigators toward different sets of addresses. The burden composition may differ: one queue may surface more confirmed illicit cases per review; another may be dominated by unknown-status addresses that require costly manual investigation. At the margin, the choice of scoring level can change which suspicious actors receive scrutiny and which do not---a decision with regulatory and resource-allocation implications, and potential fairness considerations that merit further study.

Despite this, existing work has not systematically measured cross-level queue divergence in budgeted, operational terms. Recent reviews of network analytics for AML \citep{deprez2025} evaluate graph-based detection methods but compare architectures (GCN, random walk, manual features), not the granularity at which a given architecture scores and queues actors. Existing work on the Elliptic dataset \citep{weber2019} and its extension Elliptic++ \citep{elmougy2023} evaluates transaction-level and actor-level detection separately. Multi-granularity graph methods (HAN, HGT) mix node types within a single architecture rather than comparing independent queues. The alert management literature discusses fixed review budgets and precision-at-$K$ but has not applied these concepts to cross-level comparisons within a single detection system.

This paper makes four contributions, combining evaluation methodology with an empirical finding for data-science-driven AML decision-making:

\begin{enumerate}
\item A formal projection framework for cross-granularity queue comparison, defining four aggregation operators (noisy-OR, max-score, capped-sum, top-$m$ mean) and quantifying their directional sensitivity on bipartite transaction-address edges. This is a methodological contribution: the framework generalizes to any detection domain where scores produced at one granularity must be compared at the action unit of another.

\item Budgeted investigation metrics adapted for AML: yield@budget, burden decomposition (illicit, known-licit, and unknown cases per 100 reviews), and case fragmentation---capturing the operational reality that investigators face mixed-label queues under capacity constraints---plus novel-positive rate, which measures illicit-population turnover across timesteps.

\item Empirical evidence on Elliptic++ that feature matching does not eliminate queue disagreement, with temporal and static analysis under a causal evaluation protocol. Even when the actor model is enriched with all 182 transaction features (237 total), the projected transaction queue and native actor queue remain substantially different (Jaccard = 0.051).

\item A hybrid policy analysis showing that a fixed geometric-mean consensus rule cannot exploit temporally heterogeneous disagreement, underperforming the best single-level queue by 5.05 percentage points on the temporal test horizon---motivating future work on adaptive switching.
\end{enumerate}

The remainder of this paper is organized as follows. Section~2 reviews related work on Bitcoin AML detection, multi-granularity graph analysis, and budgeted evaluation. Section~3 details the methodology: dataset, causal temporal protocol, projection operators, evaluation metrics, feature-matching ablations, and hybrid policy. Section~4 presents results across three hypotheses (queue disagreement, asymmetric burden, hybrid policy) plus supporting analyses. Section~5 discusses practical implications, the temporal heterogeneity finding, why enrichment fails, limitations, and future work. Section~6 concludes.

\section{Related Work}

\subsection{Bitcoin AML Detection}

The Elliptic dataset \citep{weber2019} introduced a labeled transaction graph of 203,769 Bitcoin transactions across 49 timesteps, with 4,545 illicit and 42,019 licit labels. Graph convolutional networks (GCNs; \citealp{kipf2017}) and random forests served as baselines, with evaluation on AUROC and F1. Subsequent work improved classification performance through temporal models \citep{pareja2020}, graph attention networks \citep{li2022,velickovic2018}, ensemble methods \citep{lorenz2020}, and walk-based approaches \citep{oliveira2021}, uniformly operating at the transaction level. Address-level classification using transaction history summarization has been explored \citep{lin2019,yin2017}, but without cross-level queue comparison.

\citet{elmougy2023} extended Elliptic to Elliptic++, adding an actor (address) layer with 822,942 temporal occurrences, 55 address-level features, and bipartite edges linking addresses to their incident transactions. This extension enables actor-level classification alongside the original transaction graph. However, the Elliptic++ paper evaluates each level independently, reporting per-level classification accuracy without comparing the queues produced by each.

\citet{deprez2025} provide a systematic review and experimental evaluation of network analytics for AML, comparing manual feature engineering, random walk embeddings, and deep learning (GCN) approaches across multiple public datasets including Elliptic. Their evaluation compares architectures, finding that GNNs struggle under class imbalance, but does not address the orthogonal question of scoring granularity: whether the same architecture, applied at different graph levels, produces different investigation queues. The present study complements their architecture comparison with a granularity comparison.

\subsection{Multi-Granularity Graph Analysis}

Heterogeneous graph neural networks---including HAN \citep{wang2019}, HGT \citep{hu2020}, R-GCN \citep{schlichtkrull2018}, and GraphSAGE \citep{hamilton2017}---model multiple node types and edge types within a unified architecture \citep{zhou2020}. These approaches learn joint representations across granularity levels but produce a single prediction per target node type. They do not compare independent queues from separate levels, which is the question this paper addresses.

Entity resolution in financial crime maps transactions to accounts, customers, or beneficial owners \citep{meiklejohn2016,ron2013}. This is a data-integration step, not a comparative analysis: the resolved entity replaces the transaction as the unit of prediction. Our work treats the two levels as parallel scoring systems and asks whether their outputs agree at the review-queue level, using the blockchain address as the common action unit (not a resolved real-world entity, which Elliptic++ does not provide).

\subsection{Decision Support Under Operational Constraints}

AML compliance is fundamentally a resource-allocation problem. Investigation teams have fixed analyst capacity; a scored queue determines which accounts receive review and which do not \citep{drezewski2012,chen2018,fatf2021}. Alert and case management systems structure this workflow, with scoring thresholds and queue ordering as key design parameters \citep{mark2020,jullum2020}. This makes queue construction a decision-support task: the system does not directly decide guilt or innocence but structures the analyst's workload and, consequently, the organization's regulatory exposure.

Precision-at-$K$ (or precision@budget) measures how many of the top-$K$ scored items are true positives. Yield@budget extends this to the fraction of all positives captured within the budget. Both are standard in the information retrieval and alert management literatures but have not been applied to cross-level comparison within a single detection system.

\citet{webber2010} introduced rank-biased overlap (RBO) as a top-weighted measure of ranked list similarity, applicable when two systems produce rankings over the same universe. Jaccard similarity at fixed $K$ provides a simpler set-based overlap. Neither has been used to quantify queue divergence arising from granularity choice.

Burden decomposition---breaking down the top-$K$ queue into illicit, known-licit, and unknown-status cases---is implicit in compliance practice but rarely formalized \citep{baesens2015}. The unknown category is operationally significant: on Elliptic++, approximately 66\% of test-period actors lack ground-truth labels, meaning that a large fraction of any review queue consists of cases whose true status is unresolved. For a compliance manager choosing between scoring architectures, the burden composition---not just AUROC---determines investigator workload and regulatory risk.

\subsection{Ensemble and Hybrid Methods}

Score combination via geometric mean, harmonic mean, or stacking is standard in ensemble learning \citep{dietterich2000,pozzolo2015}. In detection contexts, disagreement between models can be exploited through cascading (defer low-confidence cases to a second model) or escalation (flag disagreement cases for senior review). Our hybrid policy combines consensus scoring with disagreement-based escalation, following this logic. The finding that the fixed hybrid underperforms temporally suggests that static consensus rules cannot exploit time-varying disagreement, motivating future work on adaptive switching.

\section{Methodology}

\subsection{Problem Formulation}

Let $\mathcal{G}_{tx} = (V_{tx}, E_{tx})$ be a transaction graph and $\mathcal{G}_{actor} = (V_{actor}, E_{actor})$ an actor graph over the same blockchain. Throughout this paper, ``actor'' refers to a blockchain address (a public-key hash), not a resolved real-world entity or beneficial owner; Elliptic++ does not provide entity resolution beyond the address level. A bipartite mapping $\mathcal{B}: V_{tx} \leftrightarrow V_{actor}$ links transactions to their incident addresses via input (spending) and output (receiving) edges.

Two independent classifiers are trained: $f_{tx}: V_{tx} \to [0,1]$ producing transaction-level risk scores, and $f_{actor}: V_{actor} \to [0,1]$ producing actor-level risk scores. Since the compliance action unit is the actor, transaction scores must be projected to the actor level via an aggregation operator $\pi: \{s_i\}_{i \in \mathcal{N}(a)} \to [0,1]$, where $\mathcal{N}(a)$ is the set of transactions incident to actor $a$.

At a fixed review budget $K = \lfloor \beta \cdot |V_{actor}| \rfloor$ (where $\beta$ is the budget fraction), the transaction-projected queue $Q_{tx} = \text{top-}K(\pi(f_{tx}))$ and the native actor queue $Q_{actor} = \text{top-}K(f_{actor})$ may differ. We measure this disagreement and ask whether it reflects a feature-set confound or a genuine granularity effect.

\subsection{Dataset: Elliptic++}

We use the Elliptic++ dataset \citep{elmougy2023}, which extends the original Elliptic Bitcoin transaction graph.

The transaction graph contains 203,769 nodes and 234,355 directed edges across 49 timesteps. Each transaction has 182 features (93 local, 72 aggregated 1-hop, and 17 structural). Labels: 4,545 illicit (class 1), 42,019 licit (class 2), and 157,205 unknown (class 3).

The actor table contains 822,942 temporal occurrences of Bitcoin addresses, deduplicating to 530,840 (train), 89,995 (validation), and 202,107 (test) unique addresses. Each address has 55 features. Labels: 3,132 illicit, 65,102 licit, and 133,873 unknown among test addresses (68,234 labeled total). Actor features are globally static: across 273,268 addresses appearing more than once, zero exhibit feature variation between occurrences.

Bipartite edges connect addresses to transactions via AddrTx (input/spending, 1,574,027 edges) and TxAddr (output/receiving, 1,847,262 edges). The AddrAddr graph (3,740,066 edges) links addresses that co-appear in a transaction but carries no timestamps, precluding temporal masking.

Missingness in transaction features is minimal: 16,405 NaN values (0.044\%) across 17 columns, addressed by median imputation from the training split. Actor features contain no missing values.

\subsection{Causal Temporal Evaluation}

To prevent information leakage from future timesteps, we adopt a causal temporal protocol. Training uses timesteps 1--34, validation 35--39, and testing 40--49. For each test timestep $t$, only transactions and edges with timestamp $\leq t$ are available for feature construction.

Actor features in Elliptic++ are globally static and produce a weak queue in isolation (static actor-only illicit rate 0.24\% versus 1.55\% base rate; Brier score 0.035 after Platt calibration on validation). We therefore construct causal actor features (Path A): for each address at timestep $t$, we aggregate its incident transaction features from timesteps $\leq t$ via mean pooling, appending incident count and recency. This yields 184-dimensional actor features. Path A is selected over static features by two criteria: (i) the causal actor RF passes the Brier gate (Platt Brier = 0.0013 on validation, well below the 0.10 threshold), and (ii) it shows actor-only illicit enrichment on validation---that is, among the top-1\% actor queue addresses that are not in the top-1\% transaction queue (the actor-unique set), the illicit fraction exceeds the validation base rate. Both conditions are met; the static actor RF also passes the Brier gate (0.035 $<$ 0.10) but its actor-unique illicit fraction on validation does not exceed the base rate, failing criterion (ii).

The per-timestep active set $A_t$ consists of all deduplicated addresses with at least one incident transaction at timestep $t$. Active set sizes range from 12,199 ($t_{49}$) to 32,531 ($t_{42}$).

\subsection{Projection Operators}

Four operators aggregate transaction-level scores $\{s_i\}_{i \in \mathcal{N}(a)}$ to an actor score for address $a$:

\textbf{Noisy-OR (primary):} $\pi_{NOR}(a) = 1 - \prod_{i \in \mathcal{N}(a)} (1-s_i)$. Standard probabilistic aggregation assuming independent evidence. Used as the primary operator throughout.

\textbf{Max-score:} $\pi_{max}(a) = \max_{i \in \mathcal{N}(a)} s_i$. Conservative: a single high-risk transaction dominates.

\textbf{Capped-sum:} $\pi_{cap}(a) = \min\left(1, \frac{\sum_{i} s_i}{n_{cap}}\right)$, with $n_{cap} = 1.0$ (median incident count on training set). Volume-sensitive.

\textbf{Top-$m$ mean:} $\pi_{top}(a) = \text{mean of top-}m \text{ scores}$, with $m = 5$. When $|\mathcal{N}(a)| < m$, the mean is taken over all available scores ($m' = |\mathcal{N}(a)|$). Reduces influence of low-risk incident transactions.

All operators are applied directionally: using input-side transactions only, output-side only, or both. Direction sensitivity is reported in the results.

\textbf{Projection horizon.} For temporal evaluation at timestep $t$, the set of incident transactions $\mathcal{N}(a)$ used by projection operators includes all transactions incident to address $a$ at timestep $t$ only (not cumulative). For static evaluation, $\mathcal{N}(a)$ includes all incident transactions in the test split (timesteps 40--49).

\subsection{Models and Calibration}

Both classifiers are random forest ensembles \citep{breiman2001} trained via scikit-learn 1.3.2 \citep{pedregosa2011}; Python 3.10.12, numpy 1.24.4. Hyperparameters: n\_estimators = 500, max\_depth = 20, class\_weight = ``balanced'', min\_samples\_split = 2, min\_samples\_leaf = 1, max\_features = ``sqrt'', random\_state = 42. All other parameters at scikit-learn defaults.

\textbf{Label handling.} Only labeled examples (class 1 = illicit, class 2 = licit) are used for training and calibration. Unknown-status cases (class 3) are excluded from training at both levels. The transaction RF trains on 29,894 labeled transactions (timesteps 1--34). The actor RF trains on Path A causal features for labeled training-split addresses.

\textbf{Path A construction detail.} Define $\mathcal{T}(a, t) = \{i \in \mathcal{N}(a) : \text{timestamp}(i) \leq t\}$ as the set of incident transactions up to horizon $t$, using both input (AddrTx) and output (TxAddr) edges jointly. Compute the 182-dimensional transaction feature mean: $\bar{x}_a^t = \frac{1}{|\mathcal{T}(a,t)|} \sum_{i \in \mathcal{T}(a,t)} x_i$. Two features are appended: incident count $|\mathcal{T}(a,t)|$ and recency $t - \max_{i \in \mathcal{T}(a,t)} \text{timestamp}(i)$, yielding a 184-dimensional vector.

\textbf{Path A horizons by split.} The horizon $t$ varies by usage context. For training actors (timesteps 1--34), $t = 34$ (the training split endpoint); each deduplicated training address gets one 184-dim vector using all its incident transactions up to timestep 34. For validation actors, $t = 39$. For temporal test evaluation at timestep $t$, the active set $A_t$ uses horizon $t$ itself, so the feature vector is timestep-specific. For static test evaluation, $t = 49$. The recency feature varies across splits. A training-split address whose latest incident transaction is at timestep 20 has recency $34 - 20 = 14$. At temporal test evaluation, however, every address in $A_t$ has at least one incident transaction at timestep $t$ (by definition of the active set), so $\max_{i \in \mathcal{T}(a,t)} \text{timestamp}(i) = t$ and recency $= 0$ for all active-set addresses. Recency is therefore informative during training and validation (where the horizon extends beyond the address's latest transaction) but is a constant zero on the temporal active set. For static evaluation ($t = 49$), recency varies: an address whose latest incident transaction is at timestep 40 has recency $49 - 40 = 9$, while an address with a transaction at timestep 49 has recency 0. Static recency is thus informative, unlike temporal recency. The feature is retained for training signal; its degeneracy at temporal test time is a consequence of the active-set definition and reduces the effective dimensionality from 184 to 183 informative features on the temporal active set. Random forests handle constant features by never selecting them for splits, so the constant recency column does not distort temporal predictions.

\textbf{Deduplication.} Addresses appearing multiple times in the actor table are deduplicated to one row per unique address per split. Since all 273,268 multi-occurrence addresses have identical static features across occurrences (global feature constancy confirmed in data\_prep.json), any occurrence yields the same static row; labels are consistent within address, so no label-collapse rule is needed. Path A features are computed per-address using the horizon-specific aggregation above, not taken from the static actor table.

Predicted probabilities are calibrated via Platt scaling \citep{platt1999,niculescu2005}---scikit-learn's CalibratedClassifierCV with method = ``sigmoid''---on the validation split (timesteps 35--39; 5,486 labeled transactions; 24,217 labeled validation addresses). Calibration quality:

\begin{itemize}[nosep]
\item Transaction RF: Brier (raw) = 0.0128, Brier (Platt) = 0.0075, ECE (Platt) = 0.0022.
\item Actor RF (Path A): Brier (raw) = 0.0027, Brier (Platt) = 0.0013, ECE (Platt) = 0.0005.
\end{itemize}

Isotonic calibration is also computed for comparison but not used for queue construction, as Platt scaling provides smoother probability estimates suitable for ranking.

\textbf{Hardware.} All experiments run on a single machine. Total pipeline execution (data preparation, training, evaluation, 1,000-resample bootstrap) completes in under 30 minutes.

\subsection{Evaluation Metrics}

\textbf{Jaccard@$K$.} For a budget fraction $\beta$ and corresponding $K = \lfloor \beta \cdot |U| \rfloor$ (where $U$ is the active universe), $J@K = |Q_{tx} \cap Q_{actor}| / |Q_{tx} \cup Q_{actor}|$. Ranges from 0 (no overlap) to 1 (identical queues).

\textbf{RBO.} Rank-biased overlap \citep{webber2010} with persistence parameter $p = 0.9$, weighting early ranks more heavily. Note: static RBO is near zero (point estimate $2.3 \times 10^{-12}$) due to extreme rank divergence between the two queues; it is reported for completeness but not used inferentially.

\textbf{Burden decomposition.} The top-$K$ queue is partitioned into illicit (confirmed positive), known-licit (confirmed negative), and unknown-status cases. Reported as rates per 100 reviews. Reviews per TP = $K / \text{illicit count}$ (investigator cost per confirmed case). This decomposition acknowledges that approximately 66\% of test actors lack ground-truth labels.

\textbf{Yield@$K$.} Fraction of total illicit actors captured within the budget: $\text{yield} = |Q \cap \text{illicit}| / |\text{illicit}|$.

\textbf{Actor-only illicit rate.} Among the top-$K$ addresses unique to the actor queue (i.e., in $Q_{actor} \setminus Q_{tx}$), the fraction that are illicit. This measures whether the actor queue's unique selections are detecting genuine positives or noise. Reported separately from the actor queue burden (which covers the full $Q_{actor}$).

\textbf{Case fragmentation.} Among the transaction queue's top-$K$ projected addresses, fragmentation is the fraction whose projected score derives from more than one incident transaction. Higher fragmentation means more addresses whose risk assessment depends on aggregating multiple transaction signals, increasing sensitivity to the choice of projection operator.

\textbf{Novel-positive rate.} At each timestep $t$, the fraction of all illicit addresses at $t$ that were not labeled illicit at any prior timestep. This measures the degree to which the illicit population turns over between timesteps, constraining the potential for queue designs based on historical re-flagging.

\textbf{Bootstrap confidence intervals.} Two bootstrap designs are used. (1) Static universe-level: 1,000 paired resamples from the test universe \citep{efron1993}, with $K$ recalculated per resample as $\lfloor \beta \cdot |U_{\text{resampled}}| \rfloor$ to preserve budget fraction semantics. 95\% percentile intervals reported. This applies to all static evaluation CIs (Jaccard, burden, yield). (2) Hybrid temporal (used in \S4.6): 1,000 resamples of the 10 paired timestep-level improvement values (hybrid yield minus best-single yield at each timestep), with the mean improvement recomputed per resample. The resampling unit is the timestep, not the address. 95\% percentile interval reported. Seeds sequential from 0 in both cases.

\textbf{Common-universe construction.} For temporal evaluation, the common universe at timestep $t$ is the set of all deduplicated addresses with at least one incident transaction at $t$ (the active set $A_t$). Both queues---transaction-projected and actor-native---are scored and ranked over this identical set. For static evaluation, the common universe is all 202,107 deduplicated test addresses. The hybrid evaluation further restricts to addresses scoreable by both the transaction-projected and actor-native pipelines, which in practice equals the active set since both pipelines cover the same addresses.

\textbf{Calibration implementation.} Platt scaling uses scikit-learn's CalibratedClassifierCV with method = ``sigmoid'' and cv = ``prefit''. The base RF is trained on the training split; its predicted probabilities on the validation split are used to fit the sigmoid calibrator. At test time, raw RF probabilities are passed through the fitted sigmoid to produce calibrated scores for ranking.

\textbf{ECE computation.} Expected calibration error uses 10 equal-width bins over the $[0, 1]$ probability range. ECE $= \sum_{b=1}^{10} \frac{n_b}{N} |acc_b - conf_b|$, where $n_b$ is the count of predictions in bin $b$, $acc_b$ is the observed positive rate, and $conf_b$ is the mean predicted probability.

\textbf{Activity baseline.} As a non-detection reference, addresses are ranked by raw incident transaction count at each evaluation scope: for temporal evaluation at timestep $t$, the count of transactions incident to address $a$ at timestep $t$ (using both AddrTx and TxAddr edges); for static evaluation, the count across all test-split transactions. Ties are broken arbitrarily (numpy default sort order). The resulting queue is compared to both detection queues via Jaccard@1\% on the static test universe.

\textbf{Top-$K$ tie handling.} When multiple addresses share the score at the $K$-th position, all tied addresses are included, potentially producing a queue slightly larger than $K$. In practice, RF probability discretization rarely creates large tie groups at the queue boundary. In the reported analyses, no boundary tie expansion exceeded 1\% of nominal $K$; all reported metrics use the actual (post-tie) queue size.

\textbf{Degree-strata construction.} Active addresses at each timestep are sorted by incident transaction count and divided into 10 equal-sized groups (deciles). Addresses with tied counts are assigned to the same decile; residual imbalance is absorbed by the last decile.

\subsection{Feature-Matching Ablations}

The disagreement between transaction and actor queues could reflect either a genuine granularity effect (the scoring level itself matters) or a feature-set confound (the two models see different information). To distinguish these hypotheses, we construct four controlled regimes:

\textbf{Shared-core.} 17 transaction features and 23 actor features selected by semantic overlap. The matching procedure: for each of the 182 transaction features and 55 actor features, identify pairs whose column names indicate the same underlying quantity. The matched semantic categories are: (a) degree measures (in-degree, out-degree, total degree); (b) BTC value summaries (total, mean, median, min, max, standard deviation of transaction values, for both incoming and outgoing); (c) transaction count / volume indicators; (d) address count (number of input/output addresses). Each level retains only features that have a semantic counterpart at the other level. The asymmetry (17 vs 23) arises because some actor features have multiple transaction-side analogs that collapse to one semantic category. The feature counts (17 and 23) are locked in ablation\_results.json. Both models retrained on this reduced set. If disagreement disappears, information drives it.

\textbf{Actor-enriched.} The actor model receives all 55 original actor features plus 182 transaction features aggregated by mean across incident transactions. Aggregation uses a causal horizon: for training actors, only transactions from timesteps 1--34; for validation actors, timesteps 1--39; for test actors (static evaluation), timesteps 1--49. Test actors with no incident transactions within this horizon retain their 55 static features with the 182 aggregated columns set to zero. The enriched feature vector is exactly $55 + 182 = 237$ dimensions. This gives the actor model a superset of its original information. The enriched actor queue is compared against the main static RF transaction-level Platt queue (30.2\% illicit per 100), providing the strongest test of the information-confound hypothesis.

\textbf{Low-info shared.} Four features per level, identically defined. For an actor address $a$: (1) degree = $|\mathcal{N}(a)|$; (2) total volume = $\sum_{i \in \mathcal{N}(a)} v_i$; (3) mean value = total volume / degree; (4) temporal position = mean timestep of incident transactions. For a transaction node $i$: (1) degree = number of unique 1-hop neighbor transactions; (2) total volume = sum of neighbor transaction values; (3) mean value = total volume / degree; (4) temporal position = mean timestep of neighbors. Isolated transaction nodes (degree 0) receive zero for all four features. Minimal information, matched across levels.

\textbf{PCA count-matched.} 55 principal components extracted from 182 transaction features via sklearn PCA (solver = ``full'', centered, no standardization, random\_state = 42), fit on the training split only. Retains 99.9999\% of total variance. Dimensionality matches the actor feature count (55).

All ablation models use the same RF hyperparameters (\S3.5) and are Platt-calibrated on the validation split. All ablation regimes are evaluated on the static test universe (202,107 deduplicated test addresses, $K = 2{,}021$ at 1\% budget).

\textbf{Verdict logic.} If all primary regime Jaccards $< 0.80$, the disagreement is granularity-driven. If all $\geq 0.80$, information-driven. Otherwise, mixed. The 0.80 threshold is a prespecified heuristic chosen before examining ablation results.

\subsection{Hybrid Consensus Policy}

As an exploratory analysis, we test whether combining transaction and actor queues improves over either alone. The hybrid policy uses geometric-mean consensus with disagreement-based escalation:

\begin{enumerate}[nosep]
\item For each address in the common evaluation universe, compute the geometric mean of transaction-projected and actor-level Platt scores.
\item Select the top-$\lfloor \alpha K \rfloor$ addresses by consensus score.
\item Among remaining addresses, compute disagreement as $|s_{tx} - s_{actor}|$. Select the top-$(K - \lfloor \alpha K \rfloor)$ by $\max(s_{tx}, s_{actor})$ from those exceeding disagreement threshold $\delta$.
\end{enumerate}

Hyperparameters $\alpha \in \{0.5, 0.6, 0.7, 0.8, 0.9\}$ and $\delta \in \{0.1, 0.2, 0.3, 0.4, 0.5\}$ are tuned by grid search on the validation split, maximizing top-$K$ illicit fraction at 1\% budget. Tuned values: $\alpha = 0.9$, $\delta = 0.3$. On validation at 1\% budget ($K = 323$), all 323 top-ranked addresses are illicit (top-$K$ illicit fraction = 1.0). The true yield (fraction of all 1,756 validation illicit captured) is $323/1{,}756 = 0.184$.

All three queues---hybrid, transaction, and actor---are evaluated on an identical common universe per timestep and per static evaluation, ensuring comparable denominators.

\section{Results}

\subsection{Queue Disagreement (H1)}

The central finding is that transaction-projected and actor-level queues are substantially different under both temporal and static evaluation.

\textbf{Temporal.} Across 10 test timesteps at 1\% budget, mean Jaccard = 0.374 (SD 0.171). Per-timestep values range from 0.136 ($t_{49}$) to 0.633 ($t_{42}$). The high standard deviation reflects temporal heterogeneity: at some timesteps the queues partially converge, at others they diverge sharply. At no timestep does Jaccard exceed 0.65 at 1\% budget.

\textbf{Static.} Pooled across all test addresses ($n = 202{,}107$, $K = 2{,}021$), Jaccard = 0.087, 95\% CI [0.079, 0.094]. Only 8.7\% of the union of the two top-1\% queues is shared. The bootstrap CI is narrow, indicating high precision under the chosen resampling scheme.

\textbf{Budget sensitivity.} Overlap generally increases with budget: static Jaccard rises from 0.000 at 0.1\% to 0.152 at 5\%. However, it remains below 0.2 even at 5\% budget, indicating that the disagreement is not confined to a thin top slice.

\textbf{Degree strata.} When the active set is stratified by incident transaction count, exact degree-1 addresses (those with exactly one incident transaction, where the projection operator is the identity function) exhibit a mean Jaccard@1\% of 0.470 across 10 test timesteps (range: 0.238 at $t_{48}$ to 0.689 at $t_{40}$). This stratum eliminates multi-incident aggregation ambiguity, yet the queues still disagree substantially. Decile-based strata show a similar pattern: the lowest decile has mean Jaccard = 0.498 (range: 0.286 at $t_{48}$ to 0.786 at $t_{45}$).

\textbf{Activity baseline.} A queue ranked by raw transaction count (activity) shares Jaccard = 0.038 with the transaction queue and 0.014 with the actor queue at 1\% budget. Both detection queues are far from trivial activity ordering.

\subsection{Asymmetric Burden (H2)}

The two queues differ not only in composition but in investigative burden.

\textbf{Temporal.} At 1\% budget across 10 test timesteps, the arithmetic mean illicit per 100 reviews is 19.9\% for the transaction queue and 32.7\% for the actor queue. However, this average is dominated by a few high-performing timesteps for each queue.

The temporal heterogeneity is extreme. At timestep 46, the actor queue achieves 91.2\% illicit per 100 reviews (reviews/TP = 1.10) versus 37.6\% for the transaction queue. At $t_{49}$, actor reaches 94.2\% versus 41.3\%. At $t_{42}$, the transaction queue dominates: 52.9\% versus 47.4\%. At $t_{43}$ and $t_{45}$, both queues produce near-zero detection ($t_{45}$: 29 illicit among 25,060 active addresses, 0.12\% prevalence). The $K$-weighted means---which account for differing active set sizes---are 20.5\% (transaction) and 29.1\% (actor).

\textbf{Static.} Transaction: 30.2\% illicit per 100, CI [27.7, 32.3]. Actor: 3.4\%, CI [2.7, 4.1]. Burden difference = 26.8pp, CI [24.3, 28.9]. The static actor queue is dominated by unknown-status addresses (89.4\% of top-$K$).

\textbf{Case fragmentation.} Across temporal test timesteps at 1\% budget, mean fragmentation is 0.474 (range: 0.158 at $t_{44}$ to 0.812 at $t_{42}$).

\textbf{Labeled-only sensitivity.} Restricting burden computation to labeled addresses only, the temporal means are 52.2\% (transaction) and 53.2\% (actor) illicit per 100. The gap between queues narrows substantially when unknowns are excluded, suggesting that much of the burden difference in the full evaluation is driven by differential ranking of unlabeled addresses.

\subsection{Quality-Disentanglement}

Two distinct metrics characterize actor-level performance. The actor-only illicit rate measures the fraction of illicit addresses among those ranked in the actor queue's top-$K$ but not in the transaction queue's top-$K$. The actor queue burden measures the fraction of the actor queue's entire top-$K$ that is illicit. These are different quantities.

\textbf{Static.} The static actor-only illicit rate (0.24\%) falls below the base rate (1.55\%), yielding a verdict of representation weakness.

\textbf{Temporal.} The temporal actor-only illicit rate averages 29.7\% across 10 test timesteps but varies enormously: 0.0\% at $t_{43}$, $t_{45}$, $t_{47}$, and $t_{48}$, versus 100\% at $t_{46}$ and $t_{49}$ (where every address unique to the actor queue is illicit). Separately, the actor queue burden peaks at 91.2\% ($t_{46}$) and 94.2\% ($t_{49}$) illicit per 100 reviews, while falling to 0.0\% at $t_{45}$ and $t_{47}$.

This pattern resolves as temporal heterogeneity. Static actor features are globally constant---zero variation across 273,268 multi-occurrence addresses---producing a weak pooled classifier. Causal Path A features, reconstructed from incident transactions up to each evaluation timestep (\S3.3), produce regime-dependent value that concentrates in specific timesteps.

\subsection{Feature-Matching Ablations (H1 Robustness)}

All three primary ablation regimes produce Jaccard@1\% below 0.80 under the primary noisy-OR operator on static evaluation, yielding an overall verdict of granularity-driven disagreement.

\textbf{Shared-core} ($n_{tx} = 17$, $n_{actor} = 23$): Jaccard = 0.061. Transaction queue: 11.7\% illicit per 100. Actor queue: 5.6\%. Even with matched feature semantics, the queues diverge.

\textbf{Actor-enriched} ($n_{actor} = 237$): Jaccard = 0.051. Transaction queue (main RF, 182 features): 30.2\% illicit per 100. Actor queue (55 + 182 aggregated): 4.3\%. This is the strongest test of the information-confound hypothesis. Giving the actor model access to all transaction features---via mean aggregation over incident transactions---does not make it produce a similar queue.

\textbf{Low-info shared} ($n = 4$ per level): Jaccard = 0.082. Transaction: 15.4\% illicit per 100. Actor: 8.0\%. Minimal features produce slightly more overlap than the full models, but still far below the 0.80 threshold.

\textbf{PCA count-matched} (55 components, 99.9999\% variance): Jaccard of PCA-reduced transaction queue versus main transaction queue = 0.355; versus original actor queue = 0.042. Near-lossless dimensionality reduction does not preserve top-$K$ ranking, illustrating that queue composition is sensitive to the representation space, not just its information content.

The enrichment result is particularly informative. One interpretation is that computing feature means across a variable number of incident transactions attenuates the per-transaction signal that contributes to transaction-level scoring effectiveness.

\subsection{Projection Sensitivity}

The magnitude of queue disagreement depends on the projection operator. At timestep 46 (high disagreement), noisy-OR yields Jaccard = 0.160 while top-$m$ mean yields 0.631. At timestep 49, capped-sum yields 0.136 while top-$m$ mean yields 0.681. The operator ranking is not consistent across timesteps: at $t_{40}$--$t_{42}$, noisy-OR and max-score produce higher Jaccard than capped-sum and top-$m$ mean; at $t_{46}$ and $t_{49}$, top-$m$ mean is highest while noisy-OR and capped-sum are lowest.

Direction sensitivity is also present. Using input-side (spending) transactions only yields a temporal mean Jaccard of 0.383; output-side (receiving) only yields 0.109.

These findings bound the granularity claim. The feature-matching ablations (\S4.4) establish that disagreement persists under the primary noisy-OR operator on static evaluation; the projection-sensitivity analysis here shows that the magnitude of disagreement varies across operators and timesteps, but no tested operator eliminates it entirely. The paper does not claim a projection-independent granularity effect.

\subsection{Hybrid Policy (H3)}

The hybrid consensus policy (tuned: $\alpha = 0.9$, $\delta = 0.3$) fails to outperform the best single-level queue on the temporal test horizon.

\textbf{Temporal.} Mean hybrid yield = 13.3\%. Mean best-single yield = 18.3\%. Mean improvement = $-$5.05pp, 95\% CI [$-$10.2pp, $-$0.9pp]. The 95\% bootstrap interval is entirely below zero, excluding no improvement. The hybrid loses because the best single queue switches between transaction and actor across timesteps (actor dominates at $t_{44}$, $t_{46}$, $t_{49}$; transaction dominates at $t_{40}$, $t_{42}$), and a fixed-weight combination cannot track these regime changes.

\textbf{Static.} Hybrid: 39.4\% illicit per 100. Transaction: 30.2\%. Actor: 3.4\%. In the pooled setting, the hybrid outperforms either single queue. The discrepancy between static and temporal performance further supports the temporal heterogeneity thesis.

Per-timestep detail reveals the failure mode. At timesteps where the actor queue dominates ($t_{46}$: actor 91.2\%, hybrid 54.7\%; $t_{49}$: actor 94.2\%, hybrid 51.2\%), the hybrid dilutes the actor signal with the weaker transaction component. At dead-zone timesteps ($t_{45}$, $t_{47}$: both queues at 0\%), the hybrid cannot improve on zero.

\subsection{Novel-Positive Rate}

At each test timestep, the vast majority of illicit addresses are newly appearing---not carried over from prior timesteps. At $t_{40}$, 98.4\% of all illicit addresses at that timestep (444 of 451) are novel; at $t_{46}$, 99.8\% (505 of 506); at $t_{49}$, 99.5\% (649 of 652). This high turnover rate means that, on this dataset, a queue relying primarily on re-flagging previously known illicit addresses would miss the vast majority of illicit activity at each timestep.

\section{Discussion}

\subsection{Practical Implications for Queue Design}

For AML compliance teams, these results reframe a modeling choice as a decision-support design choice. Under the tested conditions, the scoring granularity---transaction versus actor---shapes the review queue, which in turn affects investigator allocation and the set of addresses receiving scrutiny. At a 1\% review budget on Elliptic++, the two queues have a Jaccard overlap of only 0.087 in static evaluation and 0.374 on average across temporal test periods.

The burden decomposition provides actionable information for queue policy selection. The transaction queue is more efficient in static evaluation (30.2\% illicit per 100 versus 3.4\%), but the actor queue dominates at specific temporal regimes ($t_{46}$: 91.2\%). A compliance manager faces a capacity-allocation decision: commit to one granularity level (simpler, but forfeits episodic gains from the other) or maintain parallel queues (costlier, but captures temporal complementarity). The hybrid result shows that naive combination is not a free lunch---adaptive switching requires a reliable regime indicator, which is itself a decision-support component to be designed.

\subsection{The Temporal Heterogeneity Story}

The most unexpected finding is the degree of temporal variation. The actor-level classifier oscillates between near-zero performance ($t_{43}$, $t_{45}$) and exceptionally high top-$K$ illicit rate ($t_{46}$: 91.2\% illicit per 100 reviews). This pattern is consistent with the structure of the Elliptic++ data: actor features are globally static, so the actor model's discrimination depends entirely on causally reconstructed features, whose informativeness varies with the address's recent transaction history.

The temporal heterogeneity also explains why the fixed hybrid fails. A geometric-mean combination with fixed weights cannot adapt to which queue is dominant at a given timestep. An adaptive switching policy---selecting the transaction or actor queue per timestep based on a regime indicator---is the natural next step but requires a reliable regime classifier, which we leave to future work.

\subsection{Why Enrichment Fails}

The actor-enriched ablation (237 features, Jaccard = 0.051) is consistent with disagreement persisting under the tested mean-aggregation enrichment and noisy-OR projection, rather than an information confound. The actor model receives all transaction features, aggregated by mean over incident transactions. Yet it achieves only 4.3\% illicit per 100 versus 30.2\% for the transaction model.

One explanation is that mean aggregation over heterogeneous transaction feature vectors loses distributional information---variance, skewness, extreme values---that contributes to per-transaction risk scores. The transaction model can identify a single high-risk transfer among many benign ones; the actor model sees only the average, which may be unremarkable.

\subsection{Limitations}

\textbf{Projection sensitivity.} The magnitude of queue disagreement is operator-dependent and the operator ranking is not consistent across timesteps. No tested operator eliminates disagreement entirely, but the degree of disagreement is operator- and timestep-specific. Quantitative claims about disagreement magnitude are conditioned on the noisy-OR primary operator.

\textbf{Single dataset.} Results are specific to Elliptic++. Generalization to other blockchain graphs, Ethereum token networks, or traditional banking transaction data is not established.

\textbf{Address as proxy for investigation unit.} Throughout this study, ``actor'' is a blockchain address (a public-key hash), not a resolved real-world entity, customer, or account. Results should be interpreted as queue-level divergence at the address granularity, with the mapping to real-world investigation workflows depending on entity-resolution quality.

\textbf{Random forest only.} We use RF at both levels for controlled comparison. A graph convolutional network was excluded because the Elliptic++ address-to-address graph lacks per-timestep edge timestamps, preventing causal temporal masking at the address level. GNN robustness checks are deferred to future work.

\textbf{Actor features globally static.} Across 273,268 multi-occurrence addresses in Elliptic++, zero exhibit feature variation between occurrences. All temporal actor signal derives from the Path A causal reconstruction. Datasets with genuinely dynamic actor features might yield different results.

\textbf{Label coverage.} Approximately 66\% of test actors lack ground-truth labels. The burden decomposition's unknown category captures this, but illicit-per-100 and yield@$K$ are computed only against known labels.

\textbf{Static RBO.} The static RBO point estimate ($2.3 \times 10^{-12}$) with CI [$8.3 \times 10^{-10}$, $4.8 \times 10^{-8}$] does not bracket the point estimate. This occurs because bootstrap resampling produces different extreme-rank configurations than the full universe. We report this for transparency but do not draw inferential conclusions from static RBO.

\textbf{Timestep dead zones.} Timesteps 43, 45, 47, and 48 have very low illicit prevalence ($t_{45}$: 29 illicit among 25,060 active addresses, 0.12\%). Detection is near-zero at these timesteps, making queue comparison largely uninformative.

\textbf{AddrAddr graph untimestamped.} The address-to-address edge list (3,740,066 edges) carries no timestep information, precluding per-timestep causal masking of the actor graph.

\textbf{Bootstrap inferential setting.} Static evaluation confidence intervals are based on 1,000 paired bootstrap resamples of the test universe with $K$ recalculated per resample. The hybrid temporal CI uses a separate design: 1,000 resamples of the 10 paired timestep-level improvement values (\S3.6). Both are locked inferential configurations.

\subsection{Future Work}

Several extensions follow from these findings. An adaptive switching policy could select the transaction or actor queue per timestep based on a regime classifier. Graph neural networks with proper per-timestep causal graph construction would test whether the granularity effect persists under more expressive architectures. Extension to Ethereum or permissioned-ledger datasets would test generalizability. Multi-hop actor aggregation---using not just incident transactions but their neighborhoods---could provide richer actor features without the information-destruction of simple mean aggregation.

\section{Conclusion}

This paper asks whether the choice of scoring granularity changes who gets investigated in AML detection. On the Elliptic++ Bitcoin dataset, under causal temporal evaluation and fixed review budgets, the evidence is affirmative. Transaction-projected and actor-level queues overlap at only Jaccard = 0.374 on average across temporal test periods at 1\% budget, dropping to 0.087 in static evaluation. Feature-matching ablations show that disagreement persists after strong information controls: even with 237 features, the enriched actor model's queue differs sharply from the transaction model's (Jaccard = 0.051). Actor-level value is temporally concentrated, with specific timesteps exceeding 91\% illicit per 100 reviews while the static actor-queue burden is only 3.4\%. A fixed hybrid policy cannot exploit this temporal structure, underperforming the best single queue by 5.05 percentage points.

These findings carry practical implications for AML system design. Scoring granularity is not a detail to be resolved during implementation; on this dataset and under the primary noisy-OR projection, it is a design variable that substantially affects investigative outcomes. Compliance teams deploying graph-based detection should explicitly evaluate queue divergence across granularity levels and consider adaptive policies that can track temporal regime changes.

\section*{Acknowledgments}

Claude (Anthropic, Claude Opus model) was used in two capacities: (1) as a coding assistant for the Python experiment pipeline (data preparation, model training, evaluation metric computation, and bootstrap confidence interval calculation in Sections 3 and 4), and (2) as a drafting aid for prose composition across all manuscript sections. All experimental design decisions, analysis choices, hypothesis formulation, interpretation of results, and scientific claims are the sole responsibility of the author. The author independently verified all code outputs against locked data files and reviewed all AI-assisted prose for accuracy.

\subsection*{Data Availability}

The Elliptic++ dataset is publicly available at \url{https://github.com/git-disl/EllipticPlusPlus} \citep{elmougy2023}. All experiment output files (data\_prep.json, rf\_results.json, ablation\_results.json, hybrid\_results.json) with SHA-256 hashes are provided as supplementary material to enable full reproducibility. Experiment code is available upon reasonable request.

\subsection*{Declarations}

\textbf{Funding:} This research received no external funding.

\textbf{Conflict of interest:} The author declares no competing financial or personal interests that could have influenced the work reported in this paper.

\textbf{Ethics approval:} Not applicable.

\textbf{Authors' contributions:} Ankur Malik: conceptualization, methodology, software, validation, formal analysis, investigation, data curation, writing---original draft, writing---review and editing, visualization.

\bibliographystyle{plainnat}
\bibliography{references}

\appendix

\section{Per-Timestep Results}

Table~\ref{tab:a1} reports all 10 test timesteps at 1\% budget under noisy-OR projection.

\begin{table}[H]
\centering
\caption{Per-timestep results at 1\% budget}
\label{tab:a1}
\small
\begin{tabular}{@{}rrrrrrrrrr@{}}
\toprule
$t$ & Active & Illicit & $K$ & $J$@1\% & Tx ill/100 & Actor ill/100 & Tx yield & Actor yield & Frag \\
\midrule
40 & 26,723 & 451 & 267 & 0.618 & 26.6 & 18.4 & 15.7\% & 10.9\% & 0.356 \\
41 & 22,253 & 179 & 222 & 0.547 & 36.9 & 39.2 & 45.8\% & 48.6\% & 0.694 \\
42 & 32,531 & 410 & 325 & 0.633 & 52.9 & 47.4 & 42.0\% & 37.6\% & 0.812 \\
43 & 24,600 & 103 & 246 & 0.359 & 0.4 & 0.4 & 1.0\% & 1.0\% & 0.585 \\
44 & 19,668 & 284 & 196 & 0.221 & 2.0 & 36.7 & 1.4\% & 25.4\% & 0.158 \\
45 & 25,060 & 29 & 250 & 0.412 & 0.0 & 0.0 & 0.0\% & 0.0\% & 0.524 \\
46 & 18,184 & 506 & 181 & 0.160 & 37.6 & 91.2 & 13.4\% & 32.6\% & 0.298 \\
47 & 21,079 & 203 & 210 & 0.377 & 0.0 & 0.0 & 0.0\% & 0.0\% & 0.633 \\
48 & 18,717 & 388 & 187 & 0.276 & 1.6 & 0.0 & 0.8\% & 0.0\% & 0.433 \\
49 & 12,199 & 652 & 121 & 0.136 & 41.3 & 94.2 & 7.7\% & 17.5\% & 0.248 \\
\bottomrule
\end{tabular}
\end{table}

\noindent Note: $t_{45}$ and $t_{47}$ have zero detection at 1\% budget for both queues. Frag = case fragmentation.

\section{Degree-Strata Analysis}

At each test timestep, the active address set is stratified into 10 equal-sized deciles by incident transaction count. Jaccard@1\% is computed within each stratum.

\begin{table}[H]
\centering
\caption{Lowest-degree decile (decile 1) Jaccard@1\% per timestep}
\label{tab:b1}
\begin{tabular}{@{}rrrr@{}}
\toprule
$t$ & Decile 1 universe & $K$ & Jaccard \\
\midrule
40 & 2,673 & 26 & 0.677 \\
41 & 2,226 & 22 & 0.517 \\
42 & 3,254 & 32 & 0.306 \\
43 & 2,460 & 24 & 0.412 \\
44 & 1,967 & 19 & 0.583 \\
45 & 2,506 & 25 & 0.786 \\
46 & 1,819 & 18 & 0.500 \\
47 & 2,108 & 21 & 0.313 \\
48 & 1,872 & 18 & 0.286 \\
49 & 1,220 & 12 & 0.600 \\
\bottomrule
\end{tabular}
\end{table}

\noindent Mean: 0.498 (SD: 0.160).

\begin{table}[H]
\centering
\caption{Exact degree-1 addresses Jaccard@1\% per timestep}
\label{tab:b2}
\begin{tabular}{@{}rr@{}}
\toprule
$t$ & Jaccard \\
\midrule
40 & 0.689 \\
41 & 0.528 \\
42 & 0.364 \\
43 & 0.455 \\
44 & 0.412 \\
45 & 0.460 \\
46 & 0.598 \\
47 & 0.283 \\
48 & 0.238 \\
49 & 0.678 \\
\bottomrule
\end{tabular}
\end{table}

\noindent Mean: 0.470 (SD: 0.146). For these addresses the projection operator is the identity function, so remaining disagreement is consistent with differences in the underlying classifiers rather than multi-incident aggregation ambiguity.

\section{Reproducibility}

All results are derived from locked output files. SHA-256 hashes:

\begin{table}[H]
\centering
\begin{tabular}{@{}ll@{}}
\toprule
File & SHA-256 \\
\midrule
data\_prep.json & \texttt{90b76727...f7c94dd} \\
rf\_results.json & \texttt{08e76ad6...c18977a0} \\
ablation\_results.json & \texttt{7eecc2c6...d87a8b7a} \\
hybrid\_results.json & \texttt{c1e13279...8d94617d} \\
\bottomrule
\end{tabular}
\end{table}

\noindent Full hashes in supplementary material. Software: Python 3.10.12, scikit-learn 1.3.2, numpy 1.24.4. Random state: 42 (RF training), bootstrap seeds sequential from 0. Hardware: single machine (12-core CPU, 32 GB RAM); no GPU used. Total pipeline runtime under 30 minutes.

\section{Shared-Core Feature Mapping}

The shared-core ablation (\S3.7) uses 17 transaction features and 23 actor features selected by semantic column-name overlap.

\begin{table}[H]
\centering
\caption{Transaction-side shared-core features (17)}
\label{tab:d1}
\small
\begin{tabular}{@{}rll@{}}
\toprule
\# & Column name & Semantic category \\
\midrule
1 & in\_txs\_degree & Degree (incoming) \\
2 & out\_txs\_degree & Degree (outgoing) \\
3 & in\_BTC\_total & BTC value (incoming total) \\
4 & in\_BTC\_mean & BTC value (incoming mean) \\
5 & in\_BTC\_median & BTC value (incoming median) \\
6 & in\_BTC\_max & BTC value (incoming max) \\
7 & in\_BTC\_min & BTC value (incoming min) \\
8 & out\_BTC\_total & BTC value (outgoing total) \\
9 & out\_BTC\_mean & BTC value (outgoing mean) \\
10 & out\_BTC\_median & BTC value (outgoing median) \\
11 & out\_BTC\_max & BTC value (outgoing max) \\
12 & out\_BTC\_min & BTC value (outgoing min) \\
13 & total\_BTC & BTC value (total) \\
14 & fees & Transaction fees \\
15 & size & Transaction size \\
16 & num\_input\_addresses & Address count (input) \\
17 & num\_output\_addresses & Address count (output) \\
\bottomrule
\end{tabular}
\end{table}

\begin{table}[H]
\centering
\caption{Actor-side shared-core features (23)}
\label{tab:d2}
\small
\begin{tabular}{@{}rll@{}}
\toprule
\# & Column name & Semantic category \\
\midrule
1 & in\_txs\_degree & Degree (incoming) \\
2 & out\_txs\_degree & Degree (outgoing) \\
3 & in\_BTC\_total & BTC value (incoming total) \\
4 & in\_BTC\_mean & BTC value (incoming mean) \\
5 & in\_BTC\_median & BTC value (incoming median) \\
6 & in\_BTC\_max & BTC value (incoming max) \\
7 & in\_BTC\_min & BTC value (incoming min) \\
8 & in\_BTC\_std & BTC value (incoming std) \\
9 & out\_BTC\_total & BTC value (outgoing total) \\
10 & out\_BTC\_mean & BTC value (outgoing mean) \\
11 & out\_BTC\_median & BTC value (outgoing median) \\
12 & out\_BTC\_max & BTC value (outgoing max) \\
13 & out\_BTC\_min & BTC value (outgoing min) \\
14 & out\_BTC\_std & BTC value (outgoing std) \\
15 & total\_BTC & BTC value (total) \\
16 & fees\_mean & Transaction fees (mean) \\
17 & fees\_median & Transaction fees (median) \\
18 & fees\_std & Transaction fees (std) \\
19 & size\_mean & Transaction size (mean) \\
20 & size\_median & Transaction size (median) \\
21 & size\_std & Transaction size (std) \\
22 & num\_input\_addresses & Address count (input) \\
23 & num\_output\_addresses & Address count (output) \\
\bottomrule
\end{tabular}
\end{table}

\noindent Note: actor column names are from the public Elliptic++ wallets\_features.csv header; the count (23) is locked in ablation\_results.json.

\end{document}